\documentclass[runningheads]{llncs}
\usepackage{graphicx}
\graphicspath{ {./images/} }
\begin{document}
\title{Training Bi-Encoders for Word Sense Disambiguation}
\author{Harsh Kohli\inst{1}\orcidID{0000-0003-1431-6025}}
\authorrunning{H. Kohli}

\institute{Compass, Inc.\\
Bangalore, India\\
\email{harsh.kohli@compass.com}}

\maketitle              % typeset the header of the contribution
\begin{abstract}
Modern transformer-based neural architectures yield impressive results in nearly every NLP task and Word Sense Disambiguation, the problem of discerning the correct sense of a word in a given context, is no exception. State-of-the-art approaches in WSD today leverage lexical information along with pre-trained embeddings from these models to achieve results comparable to human inter-annotator agreement on standard evaluation benchmarks. In the same vein, we experiment with several strategies to optimize bi-encoders for this specific task and propose alternative methods of presenting lexical information to our model. Through our multi-stage pre-training and fine-tuning pipeline we further the state of the art in Word Sense Disambiguation.
\keywords{Word Sense Disambiguation  \and Embedding Optimization \and Transfer Learning.}
\end{abstract}

\section{Introduction}

A long-standing problem in NLP, Word Sense Disambiguation or WSD, has seen several varied approaches over the years. The task is to determine the correct sense in which a word has been used from a predefined set of senses for a particular word. To put it another way, we try and derive the meaning of an ambiguous word from its surrounding context. Despite recent advances in neural language models, WSD continues to be a challenging task given the large number of fine-grained word senses and limited availability of annotated training data.

While the best performing systems today leverage BERT \cite{devlin-etal-2019-bert} or similar contextual embedding architectures, they are typically augmented with structural or relational knowledge from WordNet \cite{10.1145/219717.219748} as well as other information such as gloss definitions. Our approach builds upon these ideas, and we present our take on how to best incorporate this knowledge in our training. 

We use Bi-Encoders to learn a unique representation for a sentence, with a set target word. This embedding can be used to disambiguate or classify the target word into one of its many synsets using metrics such as cosine similarity or euclidean distance. We experiment with different optimization objectives to tune our encoding architecture as well as methods of injecting useful prior knowledge into our training. Empirically, we arrive at the best combination of settings and our model, at the time of writing, achieves SOTA results on popular evaluation benchmarks. 

\section{Related Work \& Motivation}

Unsupervised approaches to WSD such as the methods proposed by \cite{Lesk1986AutomaticSD} and \cite{Moro2014EntityLM} successfully leveraged knowledge-graph information as well as synset definitions through semantic networks and similar techniques. These methods were initially popular owing to the fact that they required no annotated training corpus. Initial supervised approaches comprised of word-expert systems which involved training a dedicated classifier for each word \cite{zhong-ng-2010-makes}. Traditional supervised learning approaches were used atop features such as words within a context-window, part of speech (POS) tags etc. Given the large number of classifiers, one for each target lemma, these approaches were harder to scale. They were also difficult to adapt to lemmas not seen during training (the most frequent sense was usually picked in such cases).

The first neural WSD models including systems by \cite{kageback-salomonsson-2016-word} and \cite{raganato-etal-2017-neural} used attention-LSTM architectures coupled with additional hand-crafted features such as POS tags. More recently, GlossBERT \cite{Huang2019GlossBERTBF} proposed an approach to imbue gloss information during training by creating a pairwise BERT \cite{devlin-etal-2019-bert} classifier using context sentences and synset definitions. The structured-logits mechanism used in EWISER \cite{bevilacqua-navigli-2020-breaking} goes one step further to include relational information such as hypernymy and hyponymy. They also demonstrate the benefits of incorporating this information and their approach out-performs all other systems so far.

We build upon the general ideas presented in GlossBERT, and discuss methods to improve the model performance using relational information. Moreover, GlossBERT uses a pairwise classifier or Cross-encoder which, while performing a full self-attention over the input pairs, could potentially lead to prohibitive compute costs at inference time. This is because for a lemma with $n$ distinct senses, the model has to infer over $n$ context-gloss pairs which can be expensive especially when using large Transformer \cite{transformer} based models.

Bi-Encoders, on the other hand, offer the flexibility to pre-index embeddings corresponding to a synset and utilize libraries such as FAISS \cite{JDH17} for fast vector similarity search during inference. We experiment with several optimization strategies to learn these synset embeddings and are able to improve upon the performance of GlossBERT, leading to gains both in terms of prediction time as well as model accuracy.

\section{Datasets \& Preprocessing}

\subsection{Source Datasets}

Consistent with most recent work, we use SemCor 3.0 \cite{miller-etal-1993-semantic} as our primary training corpus. SemCor consists of 226k sense tags and forms the largest manually annotated training corpus available. In addition, \cite{bevilacqua-navigli-2020-breaking} have shown that better results can be obtained when using both tagged and untagged WordNet examples as well as glosses. While the tagged glosses are central to approach, much like GlossBERT, we also add the tagged examples corpus to our final training dataset.

In our evaluation, we use the framework described by \cite{raganato-etal-2017-word} for benchmarking on WSD. The SemEval-2007 corpus \cite{pradhan-etal-2007-semeval} is used as our dev set for tuning parameters and model checkpointing, whereas SemEval-2013 \cite{navigli-etal-2013-semeval}, SemEval-2015\cite{moro-navigli-2015-semeval}, Senseval-2 \cite{edmonds-cotton-2001-senseval}, and Senseval-3 \cite{snyder-palmer-2004-english} comprise the remainder of our evaluation sets. 

\subsection{Data Preprocessing}

\begin{figure*}
 \center
  \includegraphics[width=\textwidth]{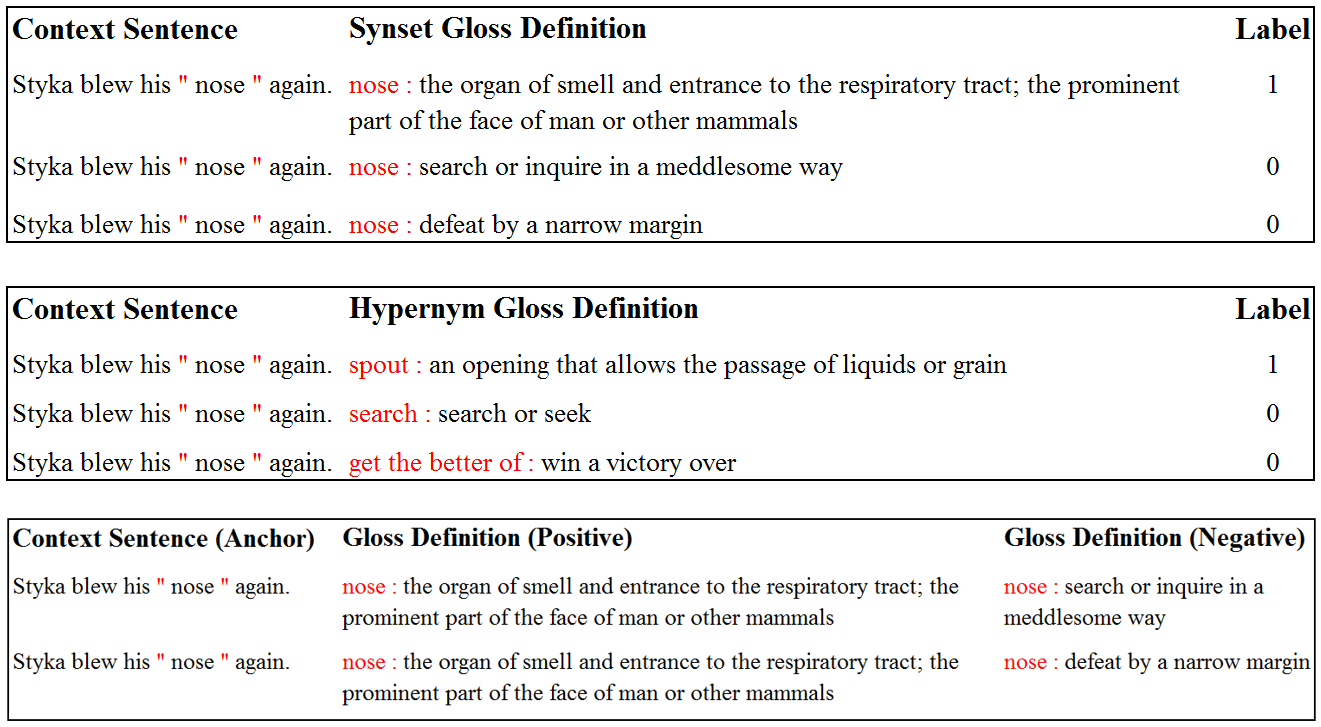}
  \caption{Context-Gloss Pairs, Context-Hypernym Pairs, and Context-Positive Gloss-Negative Gloss Triplets with Weak Supervision}
  \label{fig:sample}
\end{figure*}

We adopt the same preprocessing approach described in GlossBERT. Context sentences from SemCor as well as WordNet examples contain a signal to help identify the target word - highlighted in the context sentence column in Figure ~\ref{fig:sample}. In the gloss sentences, the lemma of the sense is prepended to the definition followed by a semicolon. This weak supervision helps better emphasize the target words in both the context as well as gloss sentences, and was also used in EWISER \cite{bevilacqua-navigli-2020-breaking} as well as MTDNN+Gloss \cite{10.1007/978-3-030-72240-1_29} in the data preprocessing step. We prepare three different datasets using the same weak supervision for our training experiments. Figure \ref{fig:sample} contains example rows corresponding to each of the datasets described in the proceeding sections.

\subsubsection{Context-Gloss Pairs}
\label{marker}

We follow GlossBERT \cite{Huang2019GlossBERTBF} for preparing our context-gloss pairwise dataset. For each target word within a context sentence, we retrieve all senses and their corresponding gloss. Thus, for a target word with $n$ different senses we have $n$ pairwise examples for training. The pair corresponding to the correct sense (gold synsets) are positive examples (label 1) while all other pairs are negative examples (label 0) in our dataset.

\subsubsection{Context-Hypernym Gloss Pairs}
\label{marker2}

For each gold (correct) synset, we take the set of immediate hypernyms. The hypernym gloss is augmented with weak signals similar to the target synset gloss. These constitute the positive pairs in our context-hypernym gloss pairs dataset. Similarly, hypernyms of incorrect synsets for a target word in a context are negative samples in this data.

\subsubsection{Context-Positive Gloss-Negative Gloss Triplets}
\label{marker3}

We use another formulation of our datasets for optimizing on the triplet loss objective \cite{7298682}. Here, the context sentence is used as an anchor, the gloss corresponding to the correct(gold) synset for the target word is the positive example, whereas gloss corresponding to each synset for the target lemma which is not the positive synset is used as a negative example. Thus, for a target word with $n$ different senses we get $n - 1$ training triplets in the anchor-positive-negative format where the anchor and positive are the same across each triplet while the negative varies. We use all the incorrect senses for each anchor-positive pair while preparing this dataset. 

\subsection{Oversampling}

Our final pairwise datasets are skewed towards the negative class as we take all the negative gloss for a target lemma, which form the negative examples in our dataset. Each context sentence, however, has only one positive example for a target word. \cite{10.1007/978-3-030-72240-1_29} augmented the positive class by generating examples using simple as well as chained back-translation. On average, 3 synthetic examples were generated for each context sentence. While we do not augment our data similarly, we use the same oversampling ratio of 3. In other words, each positive example is repeated 3 times in our training set for the pairwise training sets - both context-gloss as well as context-hypernym. The triplet loss dataset, however, is not modified as it does not suffer the same class-imbalance problem. To our context-gloss datasets, the gloss definitions of different synsets with the same lemma are added to maximize distance between gloss definitions themselves.

\section{Model Training}

We experiment with different optimization objectives for training the encoder using the datasets described in the previous section. We augment the model using a multi-stage pre-training and fine-tuning pipeline.

\subsection{Base Model}

MTDNN+Gloss \cite{10.1007/978-3-030-72240-1_29} has shown improved results over vanilla GlossBERT through a pre-training approach which used Glue data \cite{wang-etal-2018-glue}. The MT-DNN \cite{liu2019mt-dnn} architecture is used to train on each individual Glue task in a multi-task training procedure. The trained weights are then used to initialize the encoder for the fine-tuning pipeline using context-gloss pairs. To benefit from a better initial sentence representation using Glue data, we use the nli-roberta-large model from sentence-transformers \cite{reimers-2019-sentence-bert}. The sentence encoding model is a Siamese Network using the RoBERTa encoder \cite{DBLP:journals/corr/abs-1907-11692} tuned on Natural Language Inference tasks such as SNLI \cite{snli:emnlp2015} and MNLI \cite{N18-1101} which are part of the Glue benchmark. The sentence-transformers library also contains functionality for the various training experiments described in this section. We also run one trial with nli-roberta-base to justify the usage of RoBERTa large (24 layers, 1024 dimension, 335m parameters) over RoBERTa base (12 layers, 768 dimension, 110m parameters).

\subsection{Context-Gloss Training}

We use the context-gloss pairwise dataset described in section \ref{marker} to train the Bi-Encoder using various optimization strategies. Let us assume that $u$ and $v$ correspond to the pooled sentence embeddings for the context and gloss respectively, and $y \in (0, 1)$ is the label for the particular training example.

\subsubsection{Cosine Similarity Loss}

Cosine similarity between $u$ and $v$ is computed and the loss is defined as the distance between the similarity and the true label $y$.

\begin{center}
$||y - cosine\_similarity(u, v)||$
\end{center}

Here, $||.||$ corresponds to the distance metric. We use the MSE or squared error distance for our model.

\subsubsection{Contrastive Loss}

Next, we try the contrastive loss function \cite{1640964} which selectively optimizes for negative examples when they are within a certain distance or margin of each other. Mathematically, it can be defined as follows:

\begin{center}
${1\over2} (1 - y) (d(u,v))^2 + {1\over2}y(max\{0, m-(d(u,v))^2\})$
\end{center}

Here $d(u,v)$ is the cosine distance between the embeddings $u$ and $v$.

\begin{center}
    $d(u,v) = 1 - cosine\_similarity(u, v)$
\end{center}

$m$ is the margin for our contrastive loss which is set at 0.5 in our experiments.

\subsubsection{Online Contrastive Loss}

The Online Contrastive Loss defined in the sentence-transformers library is similar to the contrastive loss, except that it samples only the hard negative and positive examples from each batch. For a given batch, the maximum distance between any positive input pair ($pos\_max$, say) and minimum distance between any negative pair ($neg\_min$) is computed. Negative examples having distance less than $pos\_max$ and positive examples having distance greater than $neg\_min$ are isolated and these are treated as our hard negatives and hard positives respectively. Thereafter, contrastive loss is computed on the hard positives and negatives.

\subsection{Context-Hypernym Gloss Training}
\label{marker4}

Hypernyms to a synset correspond to a synset with a broader or more generalized meaning, of which the target synset is a subtype or hyponym. WordNet 3.0 consists of about 109k unique synsets and relational information between them such as their hypernyms, hyponyms, antonyms, entailment etc. In EWISER \cite{bevilacqua-navigli-2020-breaking}, experiments with both hypernyms and hyponyms are performed using the structured logits mechanism. Largest improvement over baselines are obtained by including hypernyms alone. The authors claim that hypernym information helps the model generalize better to synsets that are not present or under-represented in the train data (SemCor).

We only use hypernym information in our training. Immediate hypernyms for a synset are considered and the data is prepared as described in section \ref{marker2}. Pairwise models for context-hypernym pairs are used primarily in our pre-training steps. After observing results for context-gloss training we train the siamese network using the contrastive loss objective. 

\subsection{Context-Gloss \& Context-Hypernym Multi-Task Training}

Recently approaches such as MT-DNN \cite{liu2019mt-dnn} have shown strong improvement over baselines by using multi-task learning. The multi-task training procedure uses shared encoders with output layers that are specific to task types - single sentence classification, pairwise classification, pairwise similarity, and pairwise ranking. Batches from all of glue data across task types are used to train this shared encoder. Later, the encoder is individually tuned on each individual dataset to yield best results.

While MT-DNN uses a Cross-Encoder, we try and adopt the same approach in our Bi-Encoder pre-training. Context-gloss and context-hypernym pairs are treated as separate tasks and the encoder is tuned on both of these simultaneously. Similar to the context-hypernym model, the multi-task model is used as a pre-training procedure atop which further tuning is conducted. These experiments are shown in figures \ref{fig:tl1} and \ref{fig:tl2}. Like in the previous section, contrastive loss is used while training. 

\subsection{Triplet Training}

\begin{figure*}
 \center
  \includegraphics[width=\textwidth]{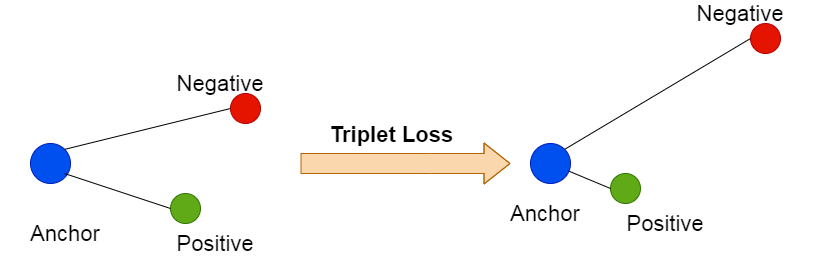}
  \caption{The triplet loss objective aims to minimize distance between the anchor and positive while maximizing the distance between anchor and negative}
  \label{fig:triplet}
\end{figure*}

Finally, we train on the triplet dataset described in section \ref{marker3}. The triplet loss \cite{7298682} uses three reference embeddings often termed the anchor, positive and negative. The loss function, as depicted in \ref{fig:triplet}, attempts to simultaneously increase the distance between the anchor and the negative, while minimizing the distance between anchor and positive. Let us consider $E_a$, $E_p$, and $E_n$ to be the embeddings for the anchor, positive, and negative samples respectively. The triplet loss is then,

\begin{center}
$max\{||E_a - E_p||^2 - ||E_a - E_n||^2 + m, 0\}$
\end{center}

Here $||.||$ denotes the euclidean distance between the embeddings. In our case, $E_a$, $E_p$, and $E_n$ correspond to embeddings for the context, gloss corresponding to the gold synset, and gloss of other synsets having the same lemma respectively. $m$ is the margin hyperparameter similar to the one in contrastive loss. We use $m = 5$ in our triplet loss training experiments.

\section{Experiments}

\begin{table*}[ht]
\small
\begin{center}
\begin{tabular}{|c|c|c|c|c|c|c|c|c|c|c|} 
\hline \bf System & \bf SE07 & \bf SE2 & \bf SE3 & \bf SE13 & \bf SE15 & \bf Noun & \bf Verb & \bf Adj & \bf Adv & \bf All\\ \hline
MFS Baseline & 54.5 & 65.6 & 66.0 & 63.8 & 67.1 & 67.7 & 49.8 & 73.1 & 80.5 & 65.5\\
Lesk\textsubscript{ext+emb} & 56.7 & 63.0 & 63.7 & 66.2 & 64.6 & 70.0 & 51.1 & 51.7 & 80.6 & 64.2\\
Babelfly & 51.6 & 67.0 & 63.5 & 66.4 & 70.3 & 68.9 & 50.7 & 73.2 & 79.8 & 66.4\\
IMS & 61.3 & 70.9 & 69.3 & 65.3 & 69.5 & 70.5 & 55.8 & 75.6 & 82.9 & 68.9\\
IMS\textsubscript{+emb} & 62.6 & 72.2 & 70.4 & 65.9 & 71.5 & 71.9 & 56.6 & 75.9 & 84.7 & 70.1\\
Bi-LSTM & - & 71.1 & 68.4 & 64.8 & 68.3 & 69.5 & 55.9 & 76.2 & 82.4 & 68.4\\
Bi-LSTM\textsubscript{+att.+LEX+POS} & 64.8 & 72.0 & 69.1 & 66.9 & 71.5 & 71.5 & 57.5 & 75.0 & 83.8 & 69.9\\
GAS\textsubscript{ext}(Linear) & - & 72.4 & 70.1 & 67.1 & 72.1 & 71.9 & 58.1 & 76.4 & 84.7 & 70.4\\
GAS\textsubscript{ext}(Concatenation) & - & 72.2 & 70.5 & 67.2 & 72.6 & 72.2 & 57.7 & 76.6 & 85.0 & 70.6\\
CAN & - & 72.2 & 70.2 & 69.1 & 72.2 & 73.5 & 56.5 & 76.6 & 80.3 & 70.9\\
HCAN & - & 72.8 & 70.3 & 68.5 & 72.8 & 72.7 & 58.2 & 77.4 & 84.1 & 71.1\\
SemCor,hyp & - & - & - & - & - & - & - & - & - & 75.6\\
SemCor,hyp(ens) & 69.5 & 77.5 & 77.4 & 76.0 & 78.3 & 79.6 & 65.9 & 79.5 & 85.5 & 76.7\\
SemCor+WNGC,hyp & - & - & - & - & - & - & - & - & - & 77.1\\
SemCor+WNGC,hyp(ens) & 73.4 & 79.7 & 77.8 & 78.7 & 82.6 & 81.4 & 68.7 & \bf 83.7 & 85.5 & 79.0\\
BERT(Token-CLS) & 61.1 & 69.7 & 69.4 & 65.8 & 69.5 & 70.5 & 57.1 & 71.6 & 83.5 & 68.6\\
GlossBERT(Sent-CLS) & 69.2 & 76.5 & 73.4 & 75.1 & 79.5 & 78.3 & 64.8 & 77.6 & 83.8 & 75.8\\
GlossBERT(Token-CLS) & 71.9 & 77.0 & 75.4 & 74.6 & 79.3 & 78.3 & 66.5 & 78.6 & 84.4 & 76.3\\
GlossBERT(Sent-CLS-WS) & 72.5 & 77.7 & 75.2 & 76.1 & 80.4 & 79.3 & 66.9 & 78.2 & 86.4 & 77.0\\
MTDNN+Gloss & 73.9 & 79.5 & 76.6 & 79.7 & 80.9 & 81.8 & 67.7 & 79.8 & 86.5 & 79.0\\
EWISER\textsubscript{hyper} & 75.2 & 80.8 & \bf 79.0 & 80.7 & 81.8 & 82.9 & 69.4 & 83.6 & 86.7 & 80.1\\
EWISER\textsubscript{hyper+hypo} & 73.8 & 80.2 & 78.5 & 80.6 & 82.3 & 82.7 & 68.5 & 82.9 & \bf 87.6 & 79.8\\
Bi-Enc\textsubscript{base} Contrastive & 75.8 & 78.3 & 76.8 & 80.1 & 81.2 & 81.6 & 69.5 & 78.5 & 84.7 & 78.6\\
Bi-Enc Cosine & 72.5 & 78.2 & 75.7 & 78.7 & 79.9 & 80.2 & 68.3 & 79.9 & 82.7 & 77.6\\
Bi-Enc Contrastive & 76.3 & 80.4 & 77.0 & 80.7 & 81.7 & 82.5 & 70.3 & 80.3 & 85.0 & 79.5\\
Bi-Enc OnlineContrastive & 75.4 & 79.4 & 78.3 & 80.3 & 82.2 & 81.8 & 71.3 & 81.1 & 85.1 & 79.4\\
Bi-Enc Multi-Task (MT) & 72.3 & 78.9 & 75.6 & 78.4 & 81.3 & 80.5 & 68.6 & 80.0 & 83.8 & 77.9\\
Bi-Enc Triplet & 75.4 & 80.4 & 77.6 & 80.7 & \bf 83.0 & 82.1 & 70.9 & 82.7 & 85.8 & 79.8\\
Bi-Enc MT+Contrastive & 76.3 & 80.1 & 77.8 & 80.6 & 81.7 & 82.3 & 70.4 & 81.3 & 86.1 & 79.6\\
Bi-Enc Hypernym+Triplet & \bf 76.7 & \bf 81.7 & 77.5 & \bf 82.4 & 82.4 & \bf 83.3 & \bf 72.0 & 80.7 & 87.0 & \bf 80.6\\
\hline
\end{tabular}
\end{center}
\caption{Final Results}
\label{table:results} 
\end{table*}

We do multiple iterations of training to empirically test the various optimization methods. We train both with and without the fine-tuning data using hypernym glosses (either directly or through multi-task pre-training). As alluded to earlier, we also train a smaller model (RoBERTa base) to observe the performance benefits of using a larger, more expressive model. At inference time, we compare the embedding of the weakly supervised context sentence with the target word against the gloss of all synsets corresponding to the target lemma. The synset with the best score (highest cosine similarity) is considered the predicted synset for the example.

Results from training the Bi-Encoder on the base model using contrastive loss, as well as larger models using cosine, contrastive, online contrastive as well as triplet loss are included in Table \ref{table:results}. We compare our results across evalution sets and POS types against recent neural approaches - Bi-LSTM \cite{kageback-salomonsson-2016-word}, Bi-LSTM + att + lex +pos \cite{raganato-etal-2017-neural}, CAN/HCAN \cite{luo-etal-2018-leveraging}, GAS \cite{luo-etal-2018-incorporating}, SemCor/SemCor+WNGC, hypernyms \cite{DBLP:journals/corr/abs-1905-05677} and GlossBERT \cite{Huang2019GlossBERTBF}, as well as knowledge based - Lesk (ext+emb) \cite{basile-etal-2014-enhanced} and Babelfly \cite{Moro2014EntityLM}, and word expert models - IMS \cite{zhong-ng-2010-makes} and IMS+emb \cite{iacobacci-etal-2016-embeddings}.

Of the large Bi-Encoder models without any transfer learning, we achieve best results using the triplet loss objective. While we do not employ any sampling heuristic, using all incorrect synset gloss for a target lemma in a context helps yield good negatives in our anchor-positive-negative formulation.

Of the pairwise approaches, we achieve best results with the contrastive loss objective. This further validates the efficacy of margin-based loss functions over traditional metrics like cosine distance for our problem. While the online contrastive loss objective often does better than simple contrastive loss, we see marginally better results with contrastive loss in this case. We reckon this is because of the inability to sample enough hard positives and negatives within a batch in a single iteration of training. While we believe that this might be mitigated and performance might improve by increasing the batch size, we are unable to test this hypothesis due to the large memory footprint of the model (335m trainable parameters) and our inability to increase batch size due to resource constraints in the form of GPU memory. However, both simple and well as online contrastive loss both comfortably outperform the cosine distance based learning objective in our experiments. Under the same settings, the larger model (RoBERTa large with contrastive loss) does better than the base model. This is in contrast to GlossBERT \cite{Huang2019GlossBERTBF}, where the authors noted better results using BERT base model over BERT large. 

\begin{figure*}
 \center
  \includegraphics[width=\textwidth]{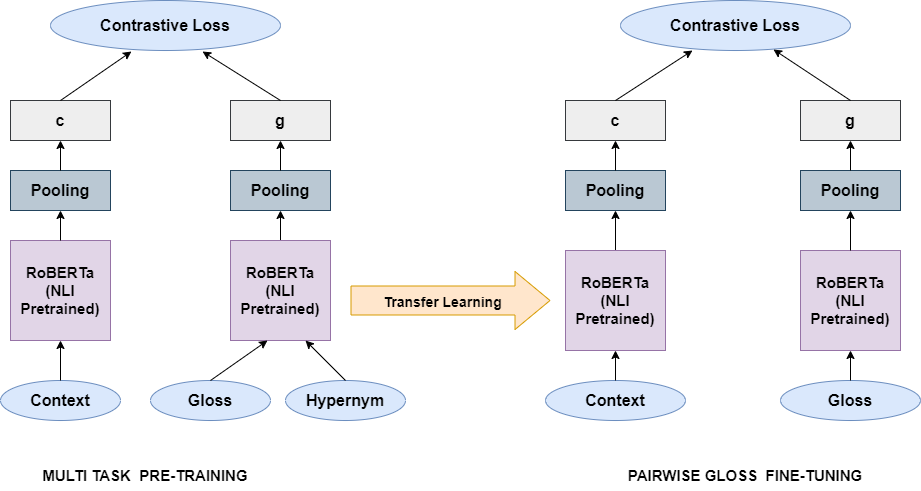}
  \caption{Transfer Learning: Multi Task Pre-training followed by Context-Gloss Fine-Tuning}
  \label{fig:tl1}
\end{figure*}

\begin{figure*}
 \center
  \includegraphics[width=\textwidth]{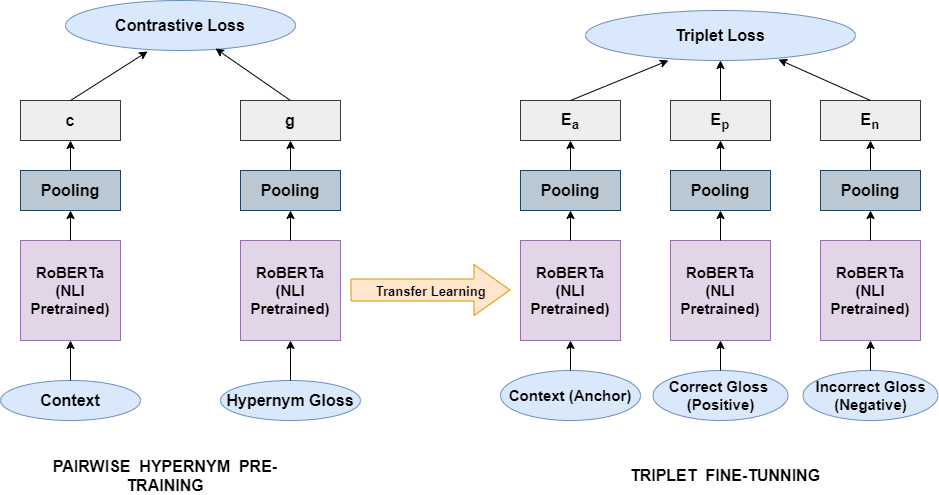}
  \caption{Transfer Learning: Context-Hypernym Pre-training followed by Triplet Objective Fine-Tuning}
  \label{fig:tl2}
\end{figure*}

Next, we try to include relational information from WordNet in our pre-training step. First, we use multi-task learning as described in section \ref{marker4} and fine-tune using context-gloss pairs and contrastive loss. We also report results of running just the multi-task model on our evaluation sets. In another experiment, we train a pairwise model using the hypernym dataset from section \ref{marker2} using contrastive loss. We fine-tune this model using triplet data from section \ref{marker3}. Architecture for the first transfer learning experiment using context-hypernym pairs and then triplet data is shown in figure \ref{fig:tl1}. The second experiment using multi-task data and context-gloss pairs is depicted in figure \ref{fig:tl2}.

Understandably, the vanilla multi-task model does not perform as well as the other pairwise Bi-Encoder models. Along with context-gloss pairs this model is also exposed to hypernym data making the training set less homogeneous and aligned with test sets. However, tuning the model with context-gloss pairs after pre-training in a multi-task setting we observe improvement in performance. This transfer learning setup, nevertheless, only gives a nominal improvement over a single-step training over context-gloss pairs.

In our other transfer learning experiment, we first train only context-hypernym pairs. Thus, we try and capture some of the inductive reasoning of hypernymy by trying to associate the target lemma in a context with its hypernym - a synset with a broader and less specific meaning. After the pre-training step we fine-tune on the triplet dataset. The pre-training step helps in this case and results in better model performance compared to the vanilla triplet training baseline. This arrangement also gives us the best results of all our experiments.

\section{Performance on Unseen Synsets}

In order to ascertain how the model performs on unseen synsets, we categorize each sample in our evaluation sets into one of two groups depending on whether the gold synset for the particular example is present in SemCor - our primary training corpus. In Table \ref{table:splits} we report results of our best models, both with and without transfer learning (TL). 

There scores reveal that there is a considerable gap in performance between the seen split (S/TL-S) and unseen split (U/TL-U). Transfer learning does help model performance on the unseen set but not as much as it does in the case of the seen split. We hypothesize that this is because the hypernym data is constructed by associating hypernyms of gold synsets with the gloss in SemCor, thus leading to limited diversity in the train set even with hypernyms outside to SemCor examples. In section \ref{conclusion}, we discuss ideas on how to further enrich the training corpus with more relational information which might help model performance on unseen synsets.

\begin{table}[h!]
\begin{center}
 \begin{tabular}{| c | c | c | c | c | c | c |} 
 \hline
\bf Test Set & \bf SE07 & \bf SE2 & \bf SE3 & \bf SE13 & \bf SE15 & \bf ALL\\ [0.5ex]  

 \hline
 U & 47.2 & 76.0 & 65.8 & 75.7 & 71.0 & 72.3\\
 S & 79.1 & 81.2 & 78.9 & 82.0 & 85.6 & 81.2\\
 TL-U & 47.2 & 74.9 & 67.3 & 77.9 & 73.4 & 72.8\\
 TL-S & 80.6 & 83.1 & 78.8 & 83.5 & 84.4 & 82.1\\
 \hline
 \end{tabular}
 \bigskip
 \caption{F1 Scores on Seen \& Unseen Splits - U - Triplet Unseen, S - Triplet Seen, TL-U - Triplet with Hypernym Pre-training Unseen, TL-S - Triplet with Hypernym Pre-training Seen}
\label{table:splits}
\end{center}
\end{table}
\vspace*{-\baselineskip}

\section{Training Parameters}

As mentioned previously, we use the sentence-transformers \cite{reimers-2019-sentence-bert} library to conduct our experiments. For every run of training, including pre-training and fine-tuning steps wherever applicable, we train for 2 epochs and checkpoint at intervals of 10000 steps. We override the previous checkpoint only if results improve on our dev set. 

A constant batch size of 32 is maintained across all our experiments. The AdamW \cite{loshchilov2018decoupled} optimizer is used with a default learning rate of $2e^{-5}$. A linear warmup scheduler is applied to slowly increase the learning rate to a constant after 10\% of training iterations (warm-up ratio = 0.1) to reduce volatility in the early iterations of training. A single Tesla V100 GPU (32 GB) is used for all training iterations.

\section{Conclusion \& Future Work}
\label{conclusion}

Through this paper, we present an approach to word sense disambiguation by training a siamese network on weakly supervised context-gloss pairs. We discuss strategies to optimize the network through various different learning objectives. We discuss methods to infuse relational information through optimizing on hypernym gloss in a distinct pre-training step. We also experiment with multi-task pre-training using hypernyms as well as gloss corresponding to the target lemma. This simple approach of pre-training with hypernyms helps us improve performance over vanilla models trained on context-gloss pairs or triplets and we achieve state of the art performance on our test sets.

While preparing our context-hypernym dataset, we only explore hypernym definitions that are one level up (one edge) from the target synset. However, it might also be worth exploring further levels to determine if pre-training with even more generalized gloss from farther synsets might help improve performance further. While hypernyms were specifically chosen as they have typically yielded the greatest improvement over baselines when included in training, we could also use additional structural features - including edges along the hyponyms, antonyms, entailment etc. In a multi-task pre-training setting, each of these datasets could be independent tasks or we could train each of them individually through a single-task, multi-stage pre-training pipeline. An ablation study might help us determine which of these features are useful. Of particular interest would be analyzing the impact of this data and additional relational information on model performance on unseen synsets.

Our best results, both with and without transfer learning, are achieved with the triplet loss objective. We employ a straight-forward strategy of creating triplets using all senses for a target lemma. However, we could also experiment with different hard or semi-hard sampling techniques \cite{7298682} to determine their impact on performance. Negatives could be sampled from the synset pool at each iteration using one of these strategies (online sampling). Finally, Poly-encoders \cite{Humeau2020Poly-encoders:} have shown promising results on several tasks. Consisting of a transformer architecture with a novel attention mechanism, they outperform Bi-Encoders purely in terms of quality. In terms of speed, they are vastly more efficient than Cross-Encoders and comparable to Bi-Encoders when running an inference against up to 1000 samples. Since we do not have as many synsets to compare against for a target lemma, Poly-encoders might be a viable alternative to the Bi-Encoders described in this paper. 

Some of the ideas discussed in this section - incorporating a richer relational information via pre-training along different edges and exploring farther beyond the immediate hypernym, better training and sampling, as well as different encoding architectures might aid in moving the needle further as we explore ways to improve performance on word sense disambiguation.

\bibliographystyle{splncs04}
\bibliography{mybibliography}
\end{document}